\documentclass[10pt,journal]{IEEEtran}%
\usepackage{amsfonts}
\usepackage{times}
\usepackage{graphicx}
\usepackage{subfigure}
\usepackage{amsmath}
\usepackage{amssymb}
\usepackage{booktabs}
\usepackage{balance}
\usepackage{hyperref}
\usepackage{caption}%
\providecommand{\U}[1]{\protect\rule{.1in}{.1in}}
\begin{document}

\title{Hierarchical Attention-based Age Estimation and Bias Analysis}
\author{Shakediel Hiba, Yosi Keller$^{\ast}$
\IEEEcompsocitemizethanks{\IEEEcompsocthanksitem S. Hiba and Y. Keller are with the Faculty of Engineering, Bar Ilan University, Ramat-Gan, Israel.\protect \and
Email: yosi.keller@gmail.com}}
\maketitle

\begin{abstract}
In this work, we present a Deep Learning approach to estimate age from facial
images. First, we introduce a novel attention-based approach to image
augmentation-aggregation, which allows multiple image augmentations to be
adaptively aggregated using a Transformer-Encoder. A hierarchical
probabilistic regression model is then proposed that combines discrete
probabilistic age estimates with an ensemble of regressors. Each regressor is
adapted and trained to refine the probability estimate over a given age range.
We show that our age estimation scheme outperforms current schemes and
provides a new state-of-the-art age estimation accuracy when applied to the
MORPH II and CACD datasets. We also present an analysis of the biases in the
results of the state-of-the-art age estimates.

\end{abstract}

\section{Introduction}

\label{sec:introduction}

Humans regularly use facial images to determine their age. There has been
significant research into accurate age estimation in computer vision and
biometrics, for a variety of applications such as e-commerce
\cite{Hakeem2012VideoAF}, face recognition \cite{Lanitis2004ComparingDC}, and the
retrieval of age-based data, to name just a few. Accurate age estimation
entails multiple computational challenges, which are uncommon for face detection or
recognition. Variations of face appearance due to aging are unknown, complex
and may be affected by multiple intrinsic and extrinsic factors, such as
ethnicity, gender, and lifestyle. Aging results in gradual changes in
appearance, making close ages appear similar, but notable age differences
alter appearance significantly \cite{Facial_Aging}.

It is common to formulate estimating age based on face images as a
classification problem, where an age $a$\ of the face $x$\ is
classified as one of $\left\{  \emph{a}_{c}\right\}  _{1}^{C}$
discrete values
\cite{eidinger2014age,guo2011simultaneous,chang2011ordinal,ZhengSun2012,ramon2012gender,guo2010human}%
, or as the regression of $\emph{a}\in\mathbb{R}^{+}$, given a
high-dimensional embedding $\widehat{\mathbf{x}}$ of the face $\emph{x}$
\cite{guo2011simultaneous,guo2010human,eidinger2014age,wang2015age,ChenGong2013,leviage}%
. The common approach to face-based biometric analysis is to first align the
face image with a canonical spatial frame \cite{RetinaFace}, and to analyze
the cropped region of interest. Early approaches utilized local image
descriptors \cite{ramon2012gender} to encode the face images in
high-dimensional representations used for regression by Kernel PLS
\cite{guo2011simultaneous}. The successful use of Deep Learning-based
approaches in a variety of computer vision tasks paved the way for the
development of end-to-end trainable age estimation schemes
\cite{8017500,deepage} using classification or regression losses. Metric
learning was used in both shallow \cite{1640964} and CNN-based
\cite{8017500,8578147,TianCCY19} schemes, where local features were learned
using their age difference as a metric measure. Ranking-based approaches
\cite{8099569,7780901, 9145576, Dark_Knowledge} apply ordinal classification
to utilize the ordinal structure of age labels to improve accuracy.

\begin{figure}[tb]
\centering%
\begin{tabular}
[c]{c}%
\subfigure[]{\includegraphics[width=0.9\linewidth]{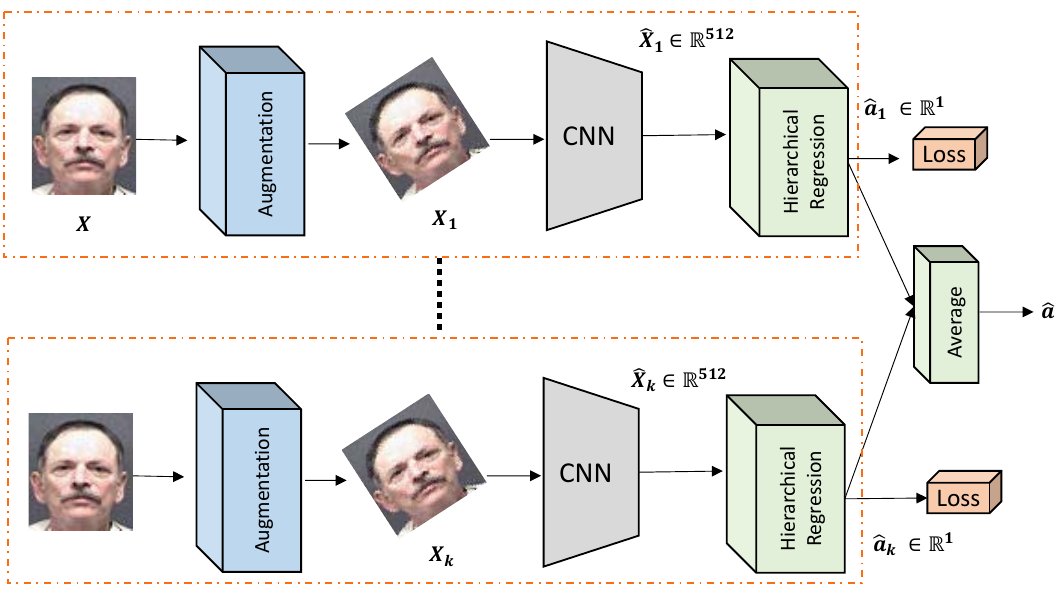}}\\
\subfigure[]{\includegraphics[width=0.9\linewidth]{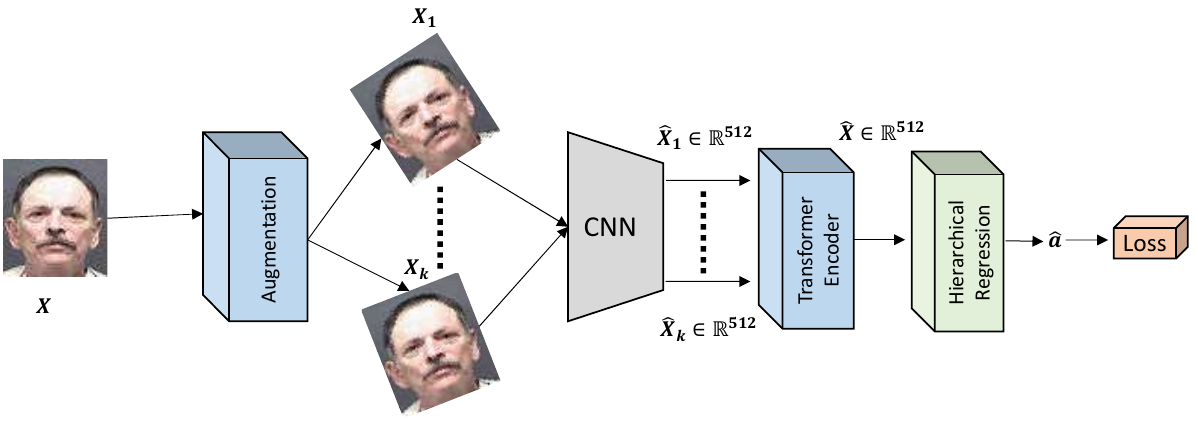}}
\end{tabular}
\caption{The proposed attention-based augmentation aggregation vs. Test Time
Augmentation (TTA) \cite{TTA-NIPS}. (a) In TTA, a network is trained using a
\textit{single} augmentation of the input $\mathbf{x}$. In \textit{run-time}
the input is augmented $\mathbf{K=10}$ times, the network is run $K$ times
\textit{separately}, and the results are averaged. (b) We propose using a \textit{ single network } to create $K=10$ different augmentations
$\{\mathbf{x}_{i}\}_{1}^{K}$ from each input $\mathbf{x}$, embed them through a
CNN backbone $\{\widehat{\mathbf{x}}_{i}\}_{1}^{K}$, and aggregate the
sequence of embeddings using the Transformer-Encoder. The aggregated feature
vector $\widehat{\mathbf{x}}$ is fed into the hierarchical regression. In
run-time the same network is run \textit{once}.}%
\label{fig:teaser}%
\end{figure}

In this work, we present a CNN-based approach to improve age estimation using
face images, which is depicted in Fig. \ref{fig:teaser}. Our \textit{core
contribution} is a novel attention-based aggregation of CNN embeddings of
multiple augmentations of each input image. The aggregation (Fig.
\ref{fig:teaser}b) is applied in each training/test iteration, allowing the
proposed network to adaptively weigh the optimal augmentations and improve the
robustness to appearance and geometrical variations. This approach follows
encoder-based schemes \cite{bert}, where a sequence of word embeddings is
aggregated in a \textit{single} embedding vector by a Transformer-Encoder
\cite{AttentionIsAllYouNeed}. In contrast, the common approach is to utilize a
\textit{single} augmentation in each training/test iteration, where multiple
augmentations are aggregated over multiple training iterations, or by using
Test Time Augmentation (TTA) \cite{TTA-NIPS} depicted in Fig. \ref{fig:teaser}%
a. Thus, our approach is applied to the embedding vectors and not to the
spatial domain of each embedding (activation map). The aggregation approach
can be applied to \textit{any} network that provides an activation map, and is
detailed and experimentally evaluated in the context of age estimation. We
also present a hierarchical probabilistic regression scheme that learns an
ensemble of age regressors, each relating to a limited age range,
respectively, and the probability of each age range. Thus, we leverage the
robustness of CNN-based probability estimation, with the precision of age
regressors of limited age domains. Our approach is shown to outperform
previous work that used classification \cite{eidinger2014age,leviage}, ordinal
classification \cite{8099569, 7780901, 9145576, coral}, tree-like structures
\cite{8954134}, and deep hierarchies \cite{8954134}.

Bias analysis is a fundamental issue in biometrics in this day and
age that was studied in the context of face recognition
\cite{9209125,Gebru,9086771,robinson2020face} and age estimation
\cite{Das_2018_ECCV_Workshops,8575487,robinson2021balancing}. We present a
bias analysis of the proposed age estimation scheme, with respect to ethnicity
and gender. Ours is the first bias analysis in face-based age estimation using
a scheme achieving state-of-the-art (SOTA) accuracy (MAE$\approx$2.5 years).

In particular, we propose the following contributions:

\begin{itemize}
\item To improve face image embedding, we propose a novel transformer-based
augmentation and embedding aggregation approach.

\item We derive a hierarchical probabilistic age estimation scheme where
the probabilistic age estimate allows us to optimally weigh the results of an
ensemble of local age regressors.

\item The proposed scheme is shown to achieve a new SOTA precision when
applied to the MORPH II \cite{1613043} and CACD \cite{chen14cross} age
estimation datasets.

\item The estimation bias of our scheme is analyzed with respect to gender and ethnicity.
\end{itemize}

\section{Related Work}

\label{sec:Related Work}

The estimation of age from a face image is challenging due to differences
between ethnicities, genders, and lifestyles in facial aging characteristics
\cite{6920084}. In their seminal work, Buolamwini and Gebru
\cite{Gebru} demonstrated the importance of ethnicity and gender in face
analysis and recognition, and this topic is of particular interest today.

Guo and Mu \cite{guo2010human} proposed a two-step procedure, which
classifies gender and ethnicity first, and then estimates age separately for
each sub group. The shallow approaches used local image features to embed
faces, followed by statistical inference. Thus, Balmaseda \emph{et al}.
\cite{ramon2012gender} used Local Binary Pattern (LBP) features and
SVM\ classifiers to compute multiscale normalized face images, alongside their
local context. Zheng and Sun \cite{ZhengSun2012} used a ranking SVM, where age
is estimated by first learning ranking relationships, which were used
alongside the reference set to estimate age. Eidinger et al.
\cite{eidinger2014age} proposed a gender and age classification scheme for
images of non-frontal faces acquired under uncontrolled conditions.
Regression-based approaches formulate the age estimation problem as a scalar
regression given a high-dimensional image embedding. Chen and Gong
\cite{ChenGong2013} introduced a cumulative attribute for learning a
regression model when only sparse and unbalanced data are available to
estimate age and crowd density. Low-level visual features extracted from
sparse and unbalanced image samples are mapped onto a cumulative attribute
space, where each dimension is related to a semantic interpretation.

CNN-based schemes forgo the use of handcrafted image descriptors, in favor of
learnt image embeddings. Thus, Wang and Kambhamettu \cite{wang2015age}
proposed a hierarchical unsupervised neural network architecture to learn
low-level translation-invariant features, used as inputs to a set of Recursive
Neural Networks (RNNs). Manifold learning was applied to capture the
underlying face aging manifold by projecting the feature vector into a
low-dimensional, better discriminative subspace. Hassner and Levi
\cite{leviage} reported significant accuracy improvements by formulating the
age estimation as a classification problem, and applying CNNs. Sendik et al.
\cite{deepage} applied deep metric learning to face features computed by a
CNN, while a Support Vector Regressor (SVR) was applied to estimate age. Deep
metric learning was also used by Liu et al. \cite{8017500}, who proposed a
hard quadruplet mining scheme to improve the resulting embedding, while a
regression-based loss was applied to estimate the age. Rothe et al.
\cite{7406390} derived a classification scheme in which the class probability
distribution of the Softmax function was used to compute the empirical
expectancy of estimated age. Pan et al. \cite{Mean-Variance} proposed a
multitask approach, first computing the empirical estimation probability of
each age using the Softmax activation function. The $L_{2}$ loss, as well as
the empirical variance of the age estimation error, were minimized. Malli et
al. \cite{Malli_2016} suggested an ensemble of CNN-based classification
models, where each ensemble model was trained to classify within a different
age domain. The final inference was computed by averaging over the models'
outputs. Tree-based approaches were also proposed \cite{8954134, 8578343}.

Shen et al. \cite{8578343} propose a novel methodology that integrates
Regression Forests with deep learning techniques. In this hybrid framework,
the nodes of the Regression Forest are engineered to adaptively partition the
input data. These nodes are then linked to fully connected layers within a
Convolutional Neural Network (CNN). The optimization of both the Random
Forests and the CNN is carried out simultaneously through an end-to-end
training process. Li et al. \cite{8954134} used a tree-based structure in
which adjacent tree leaves in nearby branches were jointly connected to create
a continuous transition, as well as an ensemble of local regressors. Each leaf
is connected to a particular local regressor. The ordinality of the estimated
ages was utilized by encoding the age labels in an ordinality-preserving
representation \cite{8099569, 7780901, 9145576, coral}, where each model
output determines whether an estimated age is higher than a given threshold.
Such approaches have been shown to improve the accuracy of age classification.
Niu et al. \cite{7780901} used an ordinal regression CNN to resolve
non-stationarity in aging patterns and developed the Asian Face Age Dataset
(AFAD) which contains more than 160K images with precise age ground-truths
labels. The Deep Cross-Population (DCP) domain adaptation approach to age
estimation was proposed by Li et al. \cite{8578147}. A CNN is trained using a
large training dataset to improve the accuracy of the age estimation on a
smaller test dataset. With DCP, transferable aging features are first learned
using the source dataset and then transferred to the target dataset. The aging
features of the two populations are then aligned using an order-preserving
pair-wise loss function. A correlation learning method for representing and
exploiting inter/intra-cumulative attribute relationships is proposed in Tian
et al. \cite{TianCCY19}. Utilizing correlations between and within gender
groups, the approach is further extended to perform gender-aware age estimations.

Multiple image datasets have been used in face-based age estimation. Some
legacy datasets such as the FG-NET \cite{cootes2008fg} (1K images), FERET
\cite{PHILLIPS1998295} (14K images), Chalearn LAP 2015
\cite{agustsson2017appareal} (7.5K images)\ and UTKFace \cite{zhifei2017cvpr}
(16K images) are too small for CNN-based schemes, while others such as the
IMDB-Wiki \cite{7406390} are based on web-scraping without an objective ground
truth age estimate, using human annotators. As the accuracy of computational
age estimation schemes improves (MAE$\approx$2.5 years), their accuracy is on
par with the human annotations, limiting their effectiveness in future works.
The MORPH Album II \cite{1613043} (60K images) is of particular importance, as
it provides accurate age labels. A particular downside of many of the
datasets, such as the AFAD \cite{7780901} and others, is that they do not
provide identity labels. This restricted some works to only using the
Random-Split (RS) test protocol, where the image set is \textit{randomly}
split into train and subsets. As in most datasets there are 5-25 images of
each subject, this inevitably leads to \textit{significant train-to-test
leakage}, making the age estimation results in these works less indicative. We
focus on datasets with identity labels, allowing us to apply the
Subject-Exclusive (SE) protocol, where \textit{all} of the images of a
particular subject are used in train \textit{or} test, \textit{but not in
both}.

\subsection{Attention and Transformers}

The Attention mechanism \cite{DBLP:journals/corr/BahdanauCB14} is a class of
contemporary neural network layers that aggregate information within input
sequences. The inputs are aggregated by computing aggregation (attention)
weights using the inner products between the input sequences. Attention layers
are often stacked to improve inference capacity, and were applied to both
sequence-to-sequence (NLP translation) and sequence-to-one (sentiment
analysis) problems. In self-attention, attention weights are calculated with
respect to a \textit{single} input series, and the module is denoted as an
\textit{Encoder}. By computing the inner products between the (single) input
sequence and itself, the encoder maps the input sequence into a higher
dimensional space. The inputs to a \textit{Decoder} are the key and query
sequences used to compute the attention weights, to aggregate the value
sequence. Attention models allow to computationally emphasize the contribution
of the task-informative image cues, in contrast to the visual clutter present
in most images. Transformers were introduced by Vaswani et al.
\cite{AttentionIsAllYouNeed} as a novel formulation of attention-based stacked
layers, allowing encoding sequences without RNN layers such as LSTM and GRU.
Transformer-based encoders and decoders utilize multiple stacked Multi-Head
Attention (MHA) and Feed Forward layers. In contrast to the sequentially
structured RNNs, the relative position and sequential order of the sequence
elements are induced by positional encodings that are added to the Attention
embeddings. Transformers were shown to provide a computationally efficient
framework for most NLP tasks \cite{bert}, achieving SOTA performance, and were
also applied in a variety of computer vision tasks \cite{DETR}. In this work,
we were inspired by recent attention-based single-sentence classification
tasks in NLP \cite{bert}, such as sentiment analysis. The gist of such
approaches is to aggregate and encode an ordered sequence of word embeddings
in a \textit{single} embedding used for inference. Multiple such NLP tasks
\cite{bert} use the same sequence (sentence) Transformer-Encoder-based
aggregation, and differ on the dataset and training loss per task. In our
approach, the sequence of augmented image embeddings is unordered. Thus, there
is no need for positional encoding.

\subsection{Bias Analysis}

\label{subsec:Related bias}

The significant accuracy improvement of face-based biometrics, such as face
recognition, gender and ethnicity identification and age estimation, has
resulted in a proliferation in their deployment in a gamut of applications. In
particular, face-based biometrics were used by law enforcement agencies
and commercial vendors. In their seminal work, Buolamwini and Gebru
\cite{Gebru} showed the inherent bias of contemporary SOTA face recognition
systems with respect to gender and ethnicity. Similarly, Wang et al.
\cite{9010843} studied ethnicity bias for multiple SOTA commercial and
academic face recognition schemes. For that, they proposed the Racial Faces in
the Wild (RFW) database, for which the ethnicity of each subject was
thoroughly validated. They also propose a domain adaptation scheme to reduce
ethnicity bias in face recognition.

Robinson et al. \cite{robinson2020face} introduced the novel Balanced Faces In
the Wild (BFW) dataset, which is gender and ethnicity-balanced, to study the
bias in state-of-the-art facial recognition systems. Different thresholds are
shown to be required for recognition across different subgroups, and there is
a significant improvement in performance. Human evaluations show that human
perception exhibits similar biases. In a subsequent study, Robinson et al.
\cite{robinson2021balancing} proposed a novel domain adaptation learning
scheme for facial embeddings computed using CNNs, to mitigate unbalanced performance between ethnicity and gender subgroups, thus improving
accuracy. The facial embeddings generated maintain identity while reducing
demographic information to enhance privacy and reduce bias.

The accuracy bias of face recognition due to ethnicity or skin tone was also
studied by Krishnapriya et al. \cite{9001031} who showed that for a fixed
decision threshold, Caucasian face images have a higher false non-matching
rate, while the face images of African-Americans are characterized by a higher
false matching rate. In particular, one-to-many identification might have a
low false-negative identification rate, while having significant
false-positive identification rates. Drozdowski et al. \cite{9086771}
presented a detailed survey of recent results in biometric algorithmic bias.
The bias in subjective age estimation by human observers was studied by Clapes
et al. \cite{8575487}. The bias is shown to be related to attributes of the
face images, such as gender, ethnicity, makeup, and expression. Additionally,
they demonstrated that using apparent age labels rather than real ages
improves the accuracy of CNN-based age estimation. The joint classification of
gender, age, and race was studied by Das et al. \cite{Das_2018_ECCV_Workshops}%
, using a multitask CNN (MTCNN), as well as the mitigation of biometric bias
on the UTKFace and Bias Estimation in Face Analytics (BEFA) datasets. Puc et
al. \cite{9287219} found that face-based age estimation is consistently more
accurate for men than for women, whereas ethnicity does not appear to have a
significant or consistent effect. Their analysis was carried out using
non-SOTA approaches, such as MAE$\approx$ 7, compared to MAE$\approx$ 2.5 in
contemporary schemes \cite{1613043}. Hence, to the best of our knowledge, our
work is the first to report SOTA age estimation with a bias analysis.

\section{Hierarchical attention-based age estimation}

\label{sec:Proposed Method}

In this work, we propose a deep learning-based scheme to estimate a subject's
age $\widehat{\emph{a}}$ given the face image $\emph{x}$. An overview of the
proposed scheme is shown in Fig. \ref{fig:teaser}. It consists of a novel
self-attention embedding (SAE) architecture based on a Transformer-Encoder,
and a hierarchical regression framework. In the image embedding phase,
detailed in Section \ref{subsec:self-atten}, given an input image $\emph{x}$,
we create $K$ corresponding augmentations $\{\emph{x}_{i}\}_{1}^{K}$, and
compute their embeddings $\mathbf{\{\hat{x}_{i}\}}_{1}^{K}$ using a backbone
CNN. The embeddings $\mathbf{\{\hat{x}_{i}\}}_{1}^{K}$ are aggregated by a
Transformer-Encoder, resulting in a fused embedding vector $\mathbf{\hat{x}}$.
The fused embedding is processed by a hierarchical regression that estimates
age probabilities, and a corresponding ensemble of age regressors, such that
the regressors' output is adaptively weighted by the classification
probabilities, as detailed in Section \ref{subsec:regression}.

\subsection{Self-Attention-based Image Embedding}

\label{subsec:self-atten}

We propose a novel augmentation and self-attention-based embedding (SAE) to
compute an image embedding that is robust to appearance variations, as shown
in Fig. \ref{fig:teaser}. For that, each input image $\emph{x}$ is augmented
$K$ times to create the set of augmented images $\{\emph{x}\boldsymbol{_{i}%
}\}_{1}^{K}\in\mathbb{R}^{224\times224}$. The image augmentations used follow
previous work and are detailed in Section \ref{subsec:Implementation Details}.
We also experimented in learning the augmentations using RandAugment
\cite{Randaugment}, but this did not improve the accuracy. Each of these image
augmentations $\mathbf{x_{k}}\in\{\mathbf{x_{k}}\}_{1}^{K}$ is embedded using
a CNN backbone. Any CNN can be used, and we evaluated multiple backbones, as
detailed in Section \ref{sec:Experiments}, to compare with the contemporary
schemes in which they were used. The set of embeddings $\mathbf{\{\hat{x}%
_{i}\}}_{1}^{K}\in\mathbb{R}^{512}$ is aggregated into a single embedding
vector $\mathbf{\hat{x}_{i}}$ using a Transformer-Encoder. As the sequence of
augmentation embeddings are unordered, there is no need for positional
encoding, and the encoding is derived by adding a fully learnt \textit{class
token} $\in\mathbb{R}^{512}$ to the encoded series. The aggregated
representation $\mathbf{\hat{x}}$ is the Transformer-Encoder's output
corresponding to the cls token. Our attention-based augmentation-aggregation
method improves image embedding for age estimation, as demonstrated in Section
\ref{subsec:dataset} through Tables \ref{table:ablation-aug} and
\ref{table:ablation-aug}

\subsection{Hierarchical probabilistic age regression}

\label{subsec:regression} \begin{figure*}[tbh]
\begin{center}
\centering\includegraphics[width=0.9\textwidth]{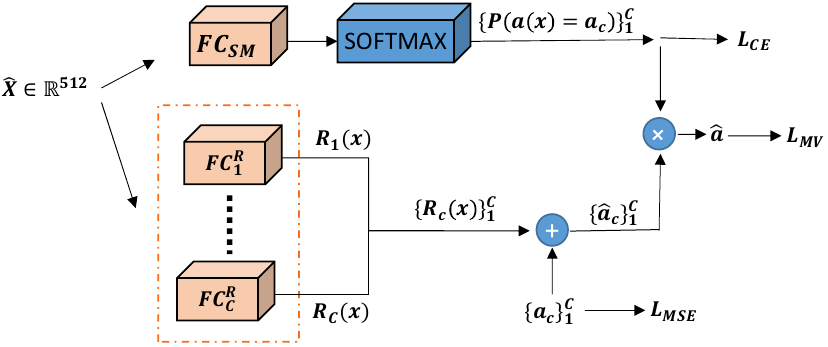}
\end{center}
\caption{The proposed hierarchical regression framework. The input feature
vector $\hat{\emph{x}}$ is jointly processed by two parallel branches: the
upper is the classifier and probability estimator, while the lower is the
regression ensemble $\left\{  R_{c}\left(  \widehat{\emph{x}}\right)
\right\}  _{1}^{C}$. The age estimate $\widehat{\emph{a}}$ is given by the
empirical expectancy of $\left\{  \widehat{\emph{a}}_{i}\right\}  _{1}^{C}$.
The upper classification subnetwork is optimized by the Cross-Entropy loss
$L_{CE}$. The outputs of the ensemble of local regressors are optimized by the
Mean Square Error loss $L_{MSE}$. The network's age estimate $\widehat
{\emph{a}}$ is optimized by the Mean-Variance Loss \cite{Mean-Variance}
$L_{MV}$. }%
\label{fig:hierarichal}%
\end{figure*}

The inference phase in deep learning-based age estimation schemes is based on
classification \cite{leviage} and regression \cite{8099569, 7780901, 9145576,
coral}. Classification-based schemes aim to classify the estimated age $a$ of
a face image into one of $\left\{  \emph{a}_{c}\right\}  _{1}^{C}$ ages. The
accuracy of regression schemes can be improved by using an ensemble of
regressors $\left\{  R_{c}\left(  \widehat{\emph{x}}\right)  \right\}
_{1}^{C}$ \cite{8954134, 8578343}, where each regressor $R_{c}\left(
\widehat{\emph{x}}\right)  $ estimates the residual regression with respect to
the discrete label $\emph{a}_{c}$. We propose to utilize the upside of both
classification and regression approaches using the framework shown in Fig.
\ref{fig:hierarichal}, where instead of classifying the age $\widehat
{\emph{a}}\in\left\{  \emph{a}_{c}\right\}  _{1}^{C}$ we estimate the age
probabilities $P\left(  \emph{a}=\emph{a}_{c}\right)  $ that are used to
estimate the age expectancy
\begin{equation}
\widehat{a}=\sum_{c}P\left(  a=a_{c}\right)  R_{c}\left(  \widehat{\mathbf{x}%
}\right)  . \label{equ:expactancy}%
\end{equation}

The classifier is optimized by multiple losses: the first is the cross-entropy
loss $L_{CE}$ that optimizes the classification probability $P\left(
\emph{a}=\emph{a}_{c}\right)  $. The second is the Mean-Variance Loss
\cite{Mean-Variance} consisting of the following two terms:
\begin{equation}
L_{M}=\frac{1}{2N}\sum_{i=1}^{N}\left(  {\sum_{c=1}^{C}}P\left(
a=a_{c}\right)  \cdot c-a_{i}^{0}\right)  ^{2} \label{equ:mean loss}%
\end{equation}

\begin{equation}
L_{V}=\frac{1}{N}\sum_{i=1}^{N}{\sum_{c=1}^{C}P\left(  a=a_{c}\right)
}\left(  {c-\sum_{c=1}^{C}c\cdot P\left(  a=a_{c}\right)  }\right)  ^{2},
\label{equ:var loss}%
\end{equation}
where $N$ is the number of points in a batch. Equation \ref{equ:mean loss}
minimizes the mean square error (MSE) between the empirical expectation and
the ground truth $\emph{a}_{i}$, while Eq. \ref{equ:var loss} minimizes the
empirical variance of the estimate. The regression ensemble $\left\{
R_{c}\left(  \widehat{\emph{x}}\right)  \right\}  _{1}^{C}$ is optimized by a
corresponding set of $L_{2}$ losses $\left\{  L_{MSE}^{c}\right\}  _{1}^{C}$,
where $L_{MSE}^{c}$ is the Mean Square Error loss applied to $\widehat{a}_{c}%
$, the result of regressor $c$. The overall loss is thus given by
\begin{equation}
L=\lambda_{1}L_{CE}+\lambda_{2}L_{M}+\lambda_{3}L_{M}+\lambda_{4}\sum
_{c}L_{MSE}^{c}, \label{eq:totatl_loss_term}%
\end{equation}
where $\left\{  \lambda_{i}\right\}  _{1}^{4}$ are predefined weights
(Section \ref{subsec:Implementation Details}).

Our work, assumes that the face aging process is episodic, implying that
although aging is a continuous process, faces from close ages are more
visually similar than others farther away, and that this process differs
notably at different ages. Hence, the age in each age episode $c$ is estimated
by a particular regressor $R_{c}\left(  \widehat{\emph{x}}\right)  $. Despite
the episodic nature of aging, it is continuous, so restricting each local
regressor $R_{c}\left(  \widehat{\emph{x}}\right)  $ may cause significant
marginal effects. Thus, we formulate the final estimation as in Eq.
\ref{equ:expactancy} to allows joint optimization and end-to-end training of
both the probability estimator and the ensemble of regressors $\left\{
R_{c}\left(  \widehat{\emph{x}}\right)  \right\}  _{1}^{C}$. Due to the
ordinal classification and the mean-variance loss, far-away local estimates
are less likely to receive a high probability, such that local estimators will
receive less significance the farther away they are from the expected age. In
case of misclassification to neighboring classes, the nearby local estimators
can compensate and provide a robust estimate.

\section{Experimental Results}

\label{sec:Experiments}

\subsection{Datasets}

\label{subsec:dataset}

MORPH Album II \cite{1613043} is one of the largest longitudinal face
databases available. It contains 55,134 facial images of 13,617 subjects with
ages 16-77, such that each subject is depicted in multiple images, the
identity of the subject in each image is known. All images are mugshots taken
in a controlled environment of good image quality, centered faces poses, and
neutral faces expressions. The dataset depicts both genders, and multiple
ethnic groups, mostly white and black. Its demographic and gender breakdown is
reported in Table \ref{table:demographic_comparison}. The CACD dataset
\cite{chen14cross} contains 163,446 images of 2,000 celebrities between the
ages of 14 and 62, collected from the Internet. The identity of the subject in
each image is given.

We follow two evaluation protocols used in previous works
\cite{Mean-Variance, 9145576} to define the training and testing
sets, The first, is the Random-Split (RS) protocol, in which the face images
are randomly split to train and test sets, such that the images of the
\textit{same} person might appear in \textit{both} train and test sets. This
creates a leakage between the train and test sets, as it essentially mixes age
estimation with age recognition. Thus, a face recognition scheme, without age
estimation training, can achieve perfect age estimation accuracy. The second
protocol is the Subject-Exclusive (SE) protocol, where identities are randomly
split to be either train or test, but not both, to avoid leakage. Due to
leakage, the RS accuracy is significantly higher than the SE scores for
\textit{all} schemes and datasets. Hence, we submit that the RS metric should
be considered less reliable and avoided in future work when possible. In this
work, we report the RS results due to legacy results that we compare to. The
MORPH II and CACD datasets can be used to assess the accuracy of age
estimation using both protocols.

We avoided using the small-scale datasets that are too small for our
Transformer-driven approach: FG-NET \cite{cootes2008fg}, FERET
\cite{PHILLIPS1998295}, Chalearn LAP 2015 \cite{agustsson2017appareal} and
UTKFace \cite{zhifei2017cvpr}. We also avoided the AFAD dataset \cite{7780901}
that is missing the identities, implying that only the RS protocol can be
applied.\begin{table}[tbh]
\caption{The demographic breakdown of the Morph II dataset \cite{1613043}.}%
\label{table:demographic_comparison}
\centering
\renewcommand{\arraystretch}{1.3}
\begin{tabular}
[c]{@{}lrrccc}%
\toprule & $\mathbf{Black}$ & $\mathbf{White}$ & $\mathbf{Asian}$ &
$\mathbf{Hispanic}$ & $\mathbf{Other}$\\
\midrule \textbf{Male} & 36,832 & 7,961 & 141 & 1,667 & 44\\
\textbf{Female} & 5,757 & 2,598 & 13 & 102 & 19\\
\textbf{Total} & 42,589 & 10,559 & 154 & 1,769 & 63\\
\bottomrule &  &  &  &  &
\end{tabular}
\end{table}

\subsection{Implementation Details}

\label{subsec:Implementation Details}

The age estimation accuracy is evaluated by the standard Mean Absolute Error
(MAE) that was used in all previous works. MAE is calculated using the mean
absolute error between the predicted age $\widehat{{\emph{a}}}{_{i}}$ and the
ground truth ${\emph{a}_{i}}$
\begin{equation}
MAE\mathbf{=}{\frac{1}{N}\sum\limits_{i}|}\widehat{{a}}{_{i}-a_{i}%
|,}\label{equ:MAEeq}%
\end{equation}
where $N$ is the number of test images. The lower the MAE the better the
accuracy. We used the CNN Vgg-16 \cite{vgg} and ResNet-34 \cite{7780459} CNN
backbones that were used in previous SOTA age estimation schemes, so that the
comparison could be based only on the proposed architecture rather than the
backbone. The proposed age estimation scheme is trained in two phases. First,
we adapt the CNN Vgg-16 \cite{vgg} and ResNet-34 \cite{7780459} ImageNet-based
backbones to face recognition using the MORPH II \cite{1613043} training set
and Arcface loss \cite{8953658}. We then train the entire proposed solution
end-to-end. The input face images were first detected, cropped, and aligned by
the RetinaFace detector \cite{RetinaFace} and then resized to a size of
224$\times$224. The proposed attention-based aggregation was implemented using
a Transformer-Encoder with four blocks, a dropout of $\emph{p}$ = 0.1, where
each block contains an MHA layer with four heads. Each input image was
augmented to $\emph{K}=10$\thinspace$\ $ images, such that multiple
augmentations were randomly applied with a probability of 0.5: horizontal
flips, color jittering, random affine transformation and randomly erasing
small parts of the image \cite{8954382}. We also applied the RandAugment
\cite{Randaugment} approach, but did not improve accuracy. The classifier in
Section \ref{subsec:regression} and the corresponding ensemble of classifiers
$\left\{  R_{c}\left(  \widehat{\emph{x}}\right)  \right\}  _{1}^{C}$ were
applied with $\left\{  \emph{a}_{c}\right\}  _{1}^{C}=\left\{
1,2,...,75\right\}  $. We used the Ranger optimizer, a combination of
Rectified Adam \cite{Liu2020OnTV} with the Lookahead technique
\cite{Lookahead}, and a Cosine Annealing learning rate decay \cite{SGDR}. All
experiments were carried out using a single NVIDIA GTX 1080 TI and the PyTorch
framework. Loss hyperparameters $\lambda_{i}$ in Eq. \ref{eq:totatl_loss_term}
are set to: {0.2, 0.05, 1, 1} accordingly, where the parameters $\lambda
_{1},\lambda_{2},\lambda_{3}$ were taken from \cite{Mean-Variance}, and
$\lambda_{4}$=$1$ is in accordance with $\lambda_{3}$.

\subsection{Results}

\label{subsec:results}

We compare our approach with multiple SOTA methods using the MORPH II
\cite{1613043} and CACD \cite{chen14cross} datasets, and the results are
reported in Tables \ref{table:Morph-II} and \ref{table:CACD}, respectively.
The results of previous schemes are quoted from their respective publications
that used the same protocol specifications as ours. Following previous works,
we randomly split the datasets to 80\% train and 20\% test in both the RS and
SE protocols. Table \ref{table:Morph-II} shows that our approach outperforms
all previous methods in the MORPH II data set using the RS and SE protocols.
The RS accuracy (MAE = 1.13) is significantly higher than the SE accuracy (MAE
= 2.53). We attribute that, as before, to leakage in the RS protocol, making
the RS results less indicative. \begin{table}[tbh]
\caption{Age estimation results evaluated using the\textbf{\ Morph-II dataset}
\ref{table:Morph-II}. We compare with previous SOTA schemes using both the RS
and SE protocols.}%
\label{table:Morph-II}
\centering
\renewcommand{\arraystretch}{1.3}
\begin{tabular}
[c]{@{}llcc}%
\toprule $\mathbf{Method}$ & $\mathbf{Backbone}$ & $\mathbf{MAE}$ &
$\mathbf{Protocol}$\\
\midrule Human \cite{6920084} &  & 6.30 & \\
Oh-rank\cite{chang2011ordinal} & AAM & 6.07 & RS\\
Ranking-CNN\cite{8099569} & ALEXNET & 2.96 & RS\\
OR-CNN\cite{7780901} & proprietary & 3.27 & RS\\
M-lsdml\cite{8017500} & RESNET101 & 2.89 & RS\\
Coral\cite{coral} & RESNET34 & 2.64 & RS\\
Dex\cite{7406390} & VGG16 & 3.25 & RS\\
Dex(pretrained)\cite{7406390} & VGG16 & 2.68 & RS\\
M-lsdml\cite{8017500} & VGG16 & 2.91 & RS\\
Mean-Variance\cite{Mean-Variance} & VGG16 & 2.16 & RS\\
BridgeNet\cite{8954134} & VGG16 & 2.63 & RS\\
DCDL\cite{9541205} & VGG-16-BN & 2.45 & RS\\
Knowledge Distil\cite{Dark_Knowledge} & proprietary & 1.95 & RS\\
\textbf{ours} & VGG16 & \textbf{1.13} & RS\\
\midrule Coral\cite{coral} & RESNET34 & 3.27 & SE\\
Mean-Variance\cite{Mean-Variance} & VGG16 & 2.79 & SE\\
soft-ranking\cite{9145576} & RESNET34 & 2.83 & SE\\
soft-ranking\cite{9145576} & VGG16 & 2.71 & SE\\
DCDL\cite{9541205} & VGG-16-BN & 2.62 & SE\\
\textbf{ours} & VGG16 & \textbf{2.53} & SE\\
\bottomrule &  &  &
\end{tabular}
\end{table}

\begin{figure}[ptb]
\centering%
\begin{tabular}
[c]{c}%
\subfigure[]{\includegraphics[width=1.0\linewidth]{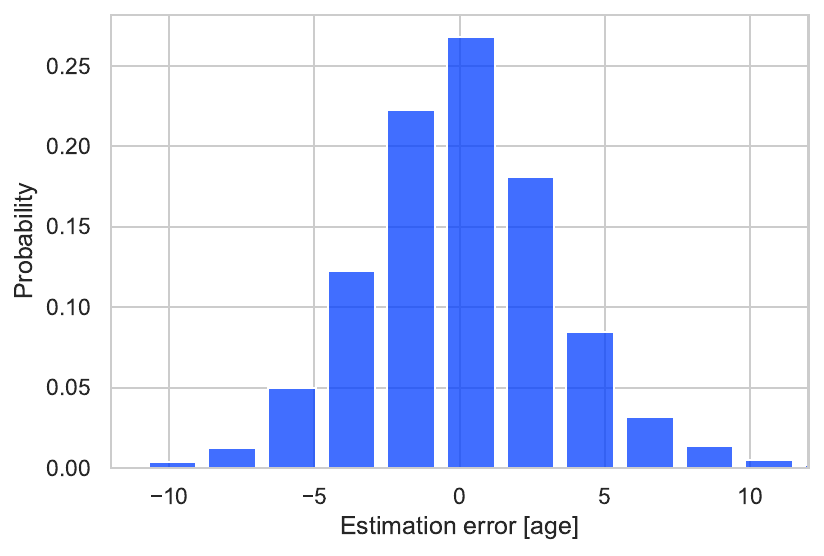}}\\
\subfigure[]{\includegraphics[width=1.0\linewidth]{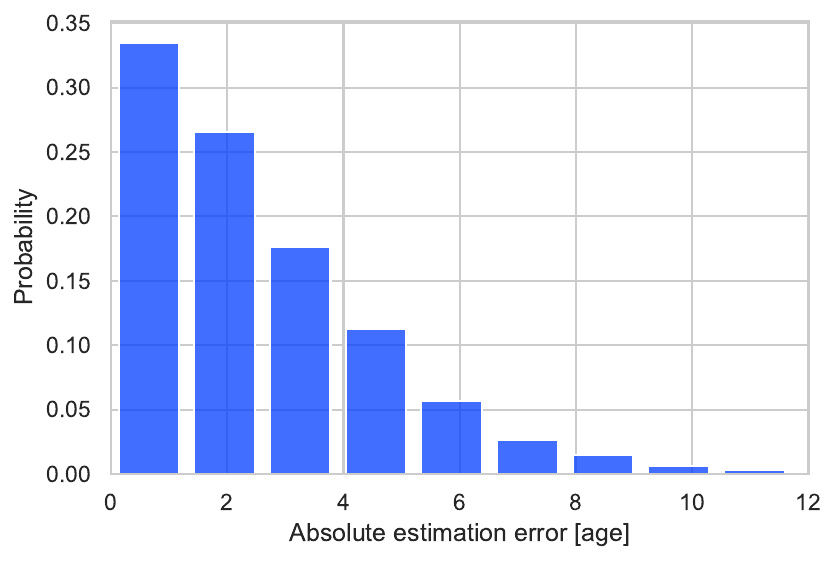}}
\end{tabular}
\caption{The distribution of the age estimation errors. The proposed scheme
was applied to the Morph II dataset.}%
\label{fig:estimation_error}%
\end{figure}

The distribution of the age estimation errors shown in Fig.
\ref{fig:estimation_error} resembles a Gaussian distribution centered around
zero, where most estimation errors ($\approx77\%$) are within an interval of
three years. In particular, our hierarchical probabilistic approach outperforms
Li et al. \cite{8954134}, that also divide the age axis into multiple
overlapping subdomains and employ local regressors over those subdomains,
while using the same backbone as ours. We also outperform the Mean-Variance
approach \cite{Mean-Variance} that employs the same backbone and
losses.\begin{table}[tbh]
\caption{Age estimation results evaluated using the \textbf{CACD dataset}
\cite{chen14cross}. We compare with previous SOTA schemes
\cite{coral,li2019facial} using the RS protocol. We also applied the SE
protocol with and without the proposed Transformer-Encoder-based augmentation
aggregation.}%
\label{table:CACD}
\centering
\renewcommand{\arraystretch}{1.3}
\begin{tabular}
[c]{@{}llcc}%
\toprule $\mathbf{Method}$ & $\mathbf{Backbone}$ & $\mathbf{MAE}$ &
$\mathbf{Protocol}$\\
\midrule Coral\cite{coral} & RESNET34 & 5.25 & RS\\
Coral\cite{coral} & Inception-v3 & 4.90 & RS\\
RNDF\cite{li2019facial} & RESNET50 & 4.60 & RS\\
\textbf{ours} & VGG16 & \textbf{4.32} & RS\\
\midrule ours(CNN) & VGG16 & 5.80 & SE\\
\textbf{ours} & VGG16 & \textbf{5.35} & SE\\
\bottomrule &  &  &
\end{tabular}
\end{table}

In Table \ref{table:CACD}, we report the results for the CACD dataset. We used
the RS protocol and compared it with the previous SOTA results of Cao et al.
\cite{coral} and Li et al. \cite{li2019facial}. Our scheme outperformed the
other schemes, although we used a shallower backbone (VGG16) compared to the
Inception-v3 \cite{coral} and RESNET50 \cite{li2019facial}. In particular, an
accuracy gain of 0.9\% was achieved compared to using the same VGG16 backbone
in \cite{coral}. We also applied the SE protocol. We are the first, to the
best of our knowledge, to evaluate it over this dataset. We compare the
variants of the proposed scheme, with and without the proposed
Transformer-Encoder augmentation aggregation. The encoder improves accuracy by
$\approx0.5\%$. As in Table \ref{table:Morph-II}, the RS errors are lower by
$\approx1\%$, implying there exists a significant leakage between the random
training and test sets.

\subsection{Ablation study}

\label{subsec:Ablation study}

To evaluate the contribution of each of the proposed algorithmic components,
we performed multiple ablation studies. In each ablation experiment, we
modified a single algorithmic component or hyperparameter to evaluate it, and
applied the resulting implementation to the MORPH II dataset that allows us to
apply both the RS and SE protocols. The baseline scheme is derived from Table
\ref{table:Morph-II}.

\begin{table}[tbh]
\caption{Ablation study of the proposed attention-based
augmentation-aggregation scheme. We compare the performance of our proposed
architecture with a different number of augmentations at input ($K$). `1
no-encoder' refers to using a single augmentation without an encoder. `10
average-pool' refers to using 10 augmentations that are aggregated by
averaging the activations maps.}%
\label{table:ablation-aug}
\centering
\renewcommand{\arraystretch}{1.3}
\begin{tabular}
[c]{@{}lc}%
\toprule $\mathbf{\# Augs}$ & $\mathbf{MAE}$\\
\midrule 1 no-encoder & 2.63\\
2 & 2.60\\
4 & 2.58\\
6 & 2.54\\
\textbf{10} & \textbf{2.53}\\
10 average-pool & 2.58\\
15 & 2.60\\
\bottomrule &
\end{tabular}
\end{table}

We first evaluated the SAE attention-based augmentation-aggregation scheme,
detailed in Section \ref{subsec:self-atten}, using the face images. For that
we implemented two additional variations: the first (1 no-encoder), is a naive
baseline without augmentations and aggregations, where we only uses a single
replica of the input image. This results in the lowest accuracy of 2.63. When
10 augmentations were applied using average pooling (10 average-pool), the
accuracy improved to 2.58 compared to our SOTA of 2.53. Regarding the number
of augmentations used, the results shown in Table \ref{table:ablation-aug}
exemplify the effectivity of the proposed augmentation-aggregation. In
particular, using additional augmentations per image improves the estimation
accuracy up to using 10 augmentations; applying additional augmentations did
not improve the accuracy. We suggest that the proposed aggregation scheme is
of general applicability and can be applied to any task where the input image
can be augmented.\begin{table}[tbh]
\caption{Ablation study of varying Transformer-Encoder parameters: the number
of encoder layers and attention heads.}%
\label{tab:encoder_size}
\centering
\strut\vspace*{-\baselineskip}
\begin{tabular}
[c]{@{}ccc}%
\toprule $\mathbf{\# layers}$ & $\mathbf{\# heads}$ & $\mathbf{MAE}$\\
\midrule 8 & 8 & 2.57\\
\textbf{4} & \textbf{4} & \textbf{2.53}\\
2 & 2 & 2.55\\
\bottomrule &  &
\end{tabular}
\end{table}

The encoder configuration was examined by trying out multiple configurations.
The more encoder layers are used, the larger the network's learning capacity.
But this might also lead to overfitting. Indeed, the results in Table
\ref{tab:encoder_size} show that the `sweet spot' is achieved for four layers
and four MHA. Using a deeper configuration leads to overfitting. We also
verified the choice of the age classification bin size and corresponding
classification ensemble $\left\{  R_{c}\left(  \widehat{\emph{x}}\right)
\right\}  _{1}^{C}$. The highest precision is achieved for $binsize=1$%
.\begin{table}[tbh]
\caption{Ablation study of the classification bin size and corresponding
regression ensemble $\left\{  R_{c}\left(  \widehat{\emph{x}}\right)
\right\}  _{1}^{C}$.}%
\label{tab:bin_size}
\centering
\renewcommand{\arraystretch}{1.3}
\begin{tabular}
[c]{@{}lc}%
\toprule $\mathbf{Bin size}$ & $\mathbf{MAE}$\\
\midrule 10 & 2.56\\
5 & 2.56\\
\textbf{1} & \textbf{2.53}\\
\bottomrule &
\end{tabular}
\end{table}

\subsection{Bias Analysis}

\label{subsec:bias}

Following the discussion in Section \ref{subsec:Related bias}, we present a
statistical bias analysis based on the MORPH II dataset, whose gender and
ethnicity breakdowns are given in Table \ref{table:demographic_comparison}.
The MORPH II dataset is imbalanced in terms of the age, gender, and ethnicity.
Our approach, as well as all prior schemes, was trained by randomly sampling
the dataset, resulting in an ethnically and gender-wise imbalanced and biased
training and test sets. We report for the first time, to the best of our
knowledge, bias estimates for a scheme that is of SOTA accuracy (MAE$\approx
$2.5 years). In our bias analysis, we use the SE protocol and the proposed
SOTA network, as in Table \ref{table:Morph-II}.

\vspace{2mm}\noindent\textbf{Age bias.} The error distribution vs. estimated
age is reported in Table \ref{table:morph2_count}. Age estimation errors
result from the given number of training samples per age range, as well as
age-related appearance changes that are hard to quantify. The 15-25 age range
has the lowest estimation error. As the number of training samples is
relatively small, we attribute it to the accelerated rate of changes in
physiological appearance, making the age estimation easier. The error flattens
for the 30-50 age range, which contains most of the training samples, but
increases for the 55-70 age range, where there are fewer
samples.\begin{table}[tbh]
\caption{\textbf{Age bias.} The number of MORPH II\ samples per age category,
and the corresponding MAE and standard deviation of the age estimation error
in each category.}%
\label{table:morph2_count}
\centering
\renewcommand{\arraystretch}{1.3}
\begin{tabular}
[c]{cccc}%
\toprule $\mathbf{Age}$ & $\mathbf{\# Samples}$ & $\mathbf{MAE}$ &
$\mathbf{Std}$\\
\midrule 15-20 & 3330 & 1.52 & 1.74\\
20-25 & 9703 & 2.00 & 1.65\\
25-30 & 8243 & 2.25 & 1.95\\
30-35 & 6243 & 2.96 & 1.99\\
35-40 & 8638 & 2.87 & 2.36\\
40-45 & 7552 & 2.69 & 1.95\\
45-50 & 5884 & 2.69 & 2.24\\
50-55 & 3421 & 3.28 & 2.47\\
55-60 & 1569 & 3.90 & 3.14\\
60-65 & 399 & 5.99 & 2.94\\
65-70 & 124 & 6.81 & 2.55\\
\bottomrule &  &  &
\end{tabular}
\end{table}

\begin{figure}[ptb]
\centering%
\begin{tabular}
[c]{c}%
\subfigure[]{\includegraphics[width=1.0\linewidth]{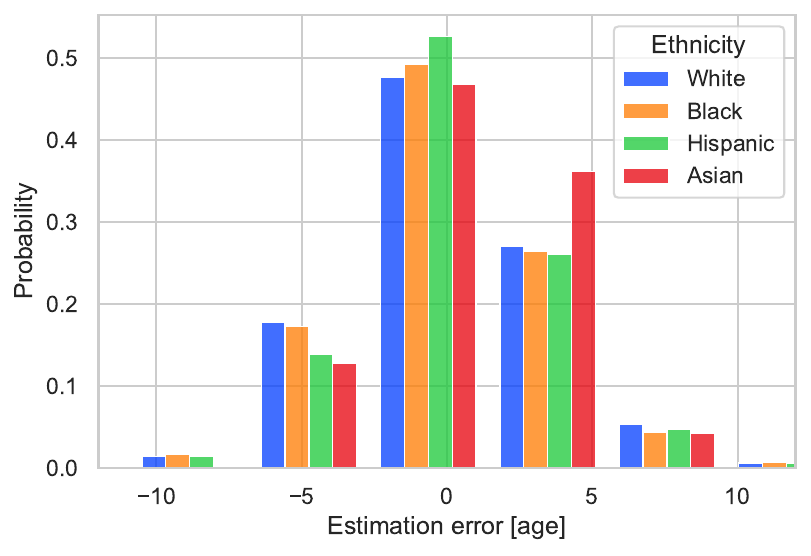}}\\
\subfigure[]{\includegraphics[width=1.0\linewidth]{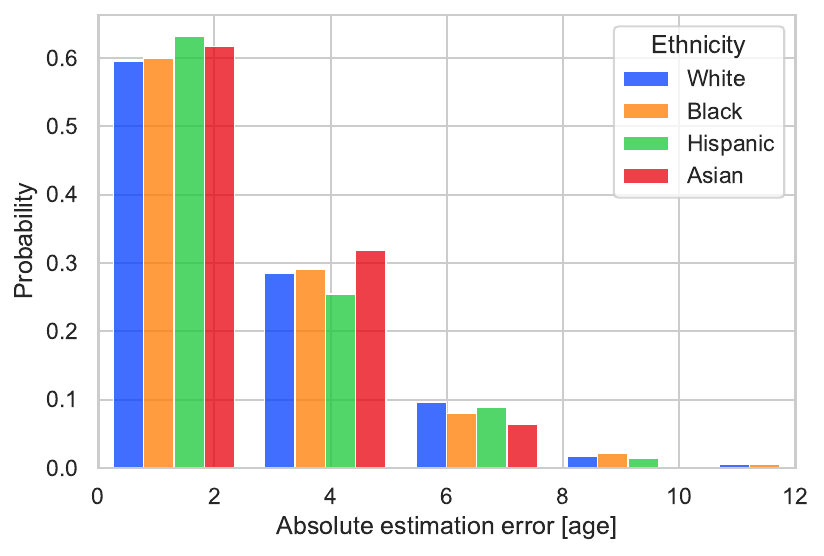}}
\end{tabular}
\caption{\textbf{Ethnicity bias.} The MAE per ethnicity over the Morph II
dataset. There is no apparent ethnicity-related estimation bias. We attribute
the slight difference in the results of Asian woman to their number (13) in
the dataset. Please consider the ethnicity/gender breakdown In Table
\ref{table:demographic_comparison}.}%
\label{fig:error_per_race}%
\end{figure}

\begin{figure}[ptb]
\centering%
\begin{tabular}
[c]{c}%
\subfigure[]{\includegraphics[width=1.0\linewidth]{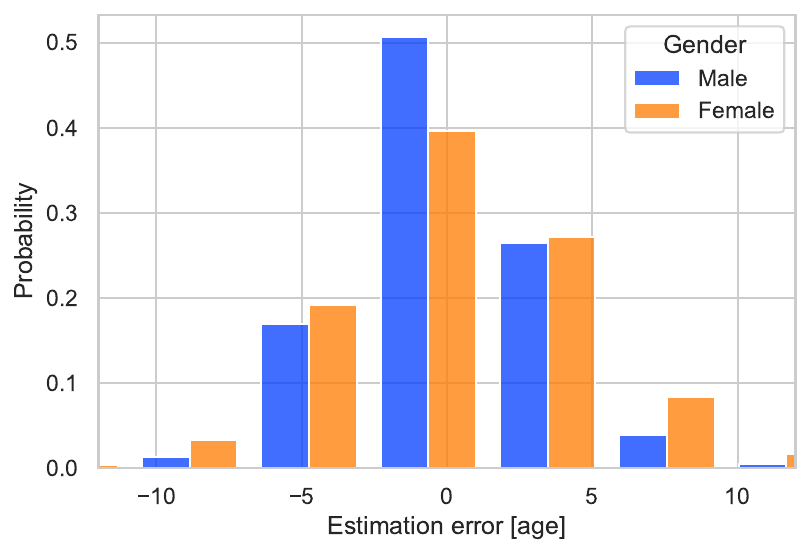}}\\
\subfigure[]{\includegraphics[width=1.0\linewidth]{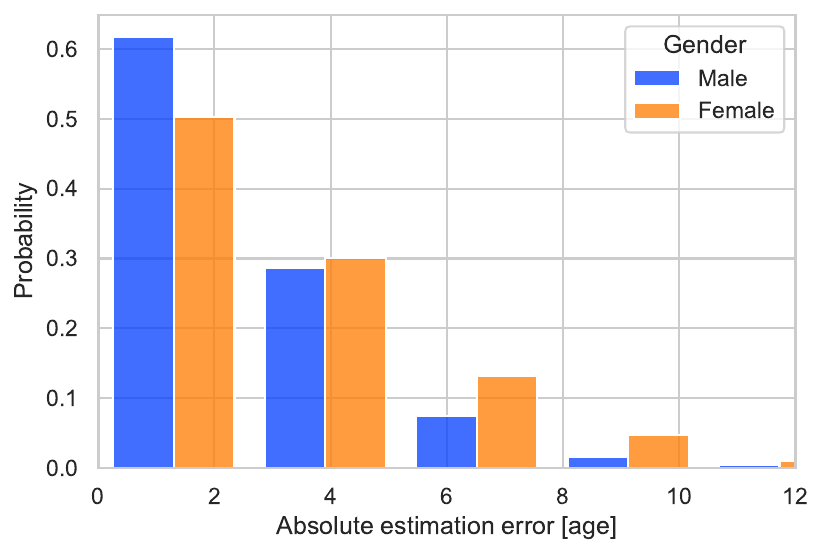}}
\end{tabular}
\caption{\textbf{Gender bias.} Estimation MAE probability per gender over the
Morph II dataset. The probability of low age estimation errors is higher by
close to 10\% for male images. }%
\label{fig:error_per_gender}%
\end{figure}

\begin{figure}[ptb]
\centering\includegraphics[width=\columnwidth,height=0.6\columnwidth]{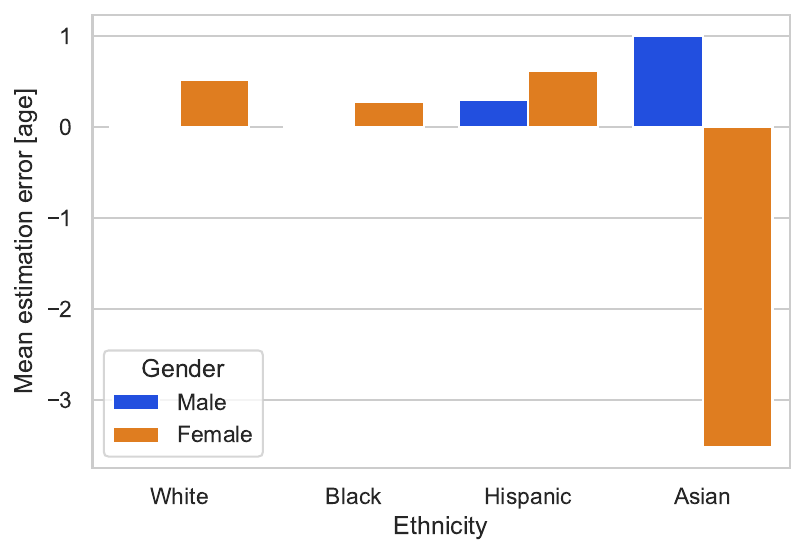}\caption{\textbf{Gender
and ethnicity bias}. The bias in age estimation per ethnicity and gender. The
age bias for men is neglible.}%
\label{fig:bias_per_gender_per_race}%
\end{figure}

\begin{figure}[ptb]
\centering\includegraphics[width=\columnwidth,height=0.6\columnwidth]{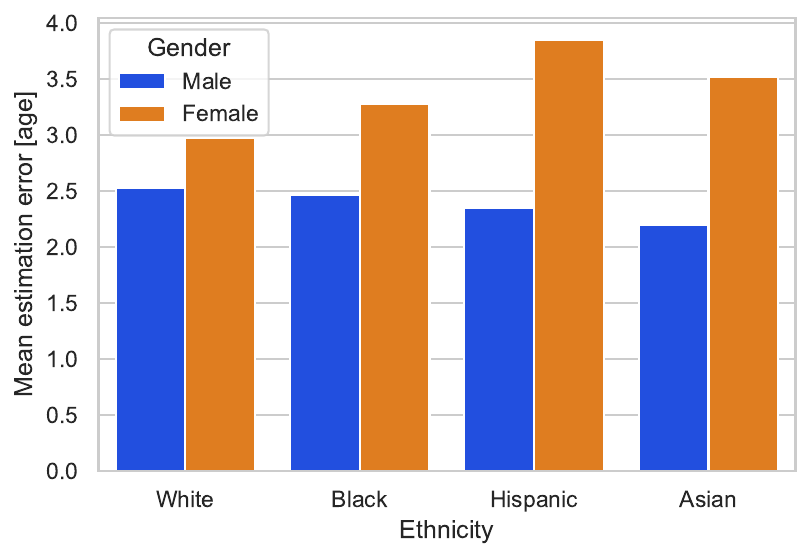}\caption{\textbf{Gender
and ethnicity MAE bias}. The MAE in age estimation per ethnicity and gender.
The MAE for all men ethnicities is similar and for each ethnicity, it
is\ smaller than that of the women.}%
\label{fig:mae_per_gender_per_race}%
\end{figure}

\vspace{2mm}\noindent\textbf{Gender and ethnicity bias.} In biometrics, gender
and ethnicity are the most common sources of estimation bias
\cite{robinson2020face}. In Figure \ref{fig:error_per_race}, we examine the
ethnicity bias, that does not appear to be significant, although the MORPH II
database is heavily skewed towards Black men, making up 67\% of the dataset
(Table \ref{table:demographic_comparison}). Figure \ref{fig:error_per_gender}
presents the gender-specific estimation accuracy. The estimation error rate
for men is lower by close to 10\% due to the large number of male training
samples. Figures \ref{fig:bias_per_gender_per_race} and
\ref{fig:mae_per_gender_per_race} breakdown the bias and MAE due to gender and
ethnicity, where the error for female subjects is greater for all ethnicities.
It is substantially higher for black and female subjects, respectively, by
nearly 0.8 and 0.5 years. However, it is biased towards female subjects by an
average error of 0.7 years. The results are similar to those of Puc et al.
\cite{9287219} that used different schemes and evaluation datasets.


\begin{figure}[ptb]
\centering\includegraphics[width=\columnwidth,height=0.6\columnwidth]{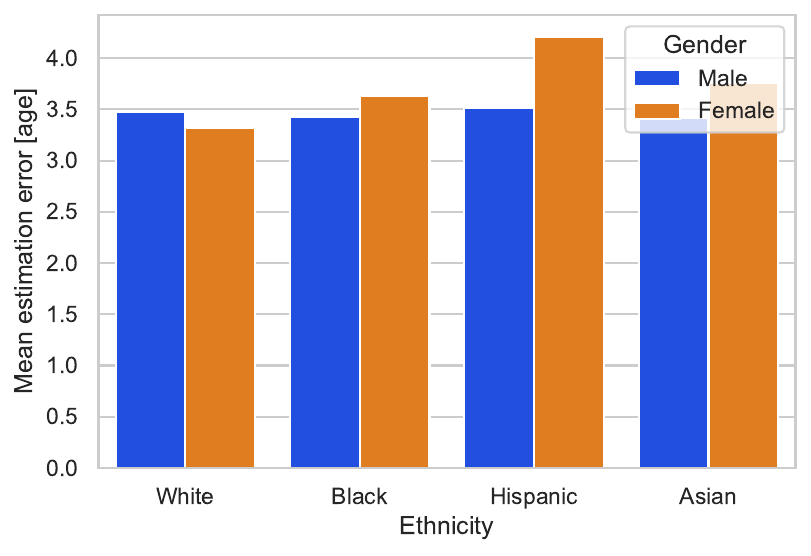}\caption{\textbf{Gender
and ethnicity MAE bias trained using the gender-balanced training set}. The
MAE of the age estimation per ethnicity and gender. The network was trained
using a gender-balanced training set based on the MORPH II dataset, reducing
the age estimation bias compared to the estimates based on the gender-skewed
MORPH II dataset in Fig. \ref{fig:mae_per_gender_per_race}.}%
\label{fig:small_mae_per_gender_per_race}%
\end{figure}

\textbf{Age estimation bias due to imbalanced training sets. }The demographics
of the MORPH II \cite{1613043} dataset, detailed in Table
\ref{table:demographic_comparison}, are significantly skewed in favor of men
vs. women, 77\% and 23\%, respectively. We mitigated the gender imbalance to
evaluate its influence. Hence, we followed Robinson et al.
\cite{robinson2021balancing,robinson2020face} by creating a gender-balanced
image set based on MORPH II dataset. We used all 8489 female images and
randomly sampled the men identities until we attained the same number of men
face images. The set was split 80\% and 20\% for the training and test tasks,
respectively. We retrained the VGG16-based age estimation network using the
gender-balanced training set using the SE protocol as in Section
\ref{subsec:results}. The results are shown in Fig.
\ref{fig:small_mae_per_gender_per_race}, and we compare to Fig.
\ref{fig:mae_per_gender_per_race} where the full MORPH II set was used. The
estimation accuracy decreased similarly for black and white men by
\symbol{126}0.95 years, while that of woman decreased by \symbol{126}0.35
years. This implies that the absolute accuracy of age estimation is mostly
influenced by the number of training samples. Since the men estimation
accuracy degraded, the bias in their favor decreased.

\subsection{Self-Attention-based Embedding}

In order to exemplify the general applicability of the Self-Attention-based
Embedding (SAE) approach, introduced in Section \ref{subsec:self-atten}, it
was applied to the classification of the CIFAR-10 and CIFAR-100 datasets
\cite{CIFAR}. The CIFAR-10 dataset consists of 10 classes with 5000 training
images and 1000 testing images per class, while the CIFAR-100 dataset has 100
classes with 500 training images and 100 testing images per class. The images
in both datasets are $32\times32$ pixels and have been previously used to
evaluate learning schemes. We used the SAE approach with varying parameters to
classify images using shallow and deep neural networks and observe their
impact on the accuracy. The CIFAR-10 dataset was tested with a shallow
\footnote{https://pytorch.org/tutorials/beginner/blitz/cifar10\_tutorial.html?highlight=data\%20loader}
and the VGG16 networks, while the more challenging CIFAR-100 dataset was
tested with the VGG16, RESNET18, RESNET34 and RESNET50 networks. We also
compared to a SAE formulation based on Average Pooling, instead of a
Transformer-Encoder. All networks were trained using a batch size of 64, an
Adam optimizer with weight decay of 1e-5, and an initial learning rate of
$1e-4$ during all simulations. The learning rate was reduced by a factor of
0.2 if the loss did not decrease for more than 3 epochs. The self-attention
was implemented using a single-layer Transformer-Encoder with two heads, which
only required the addition of two fully connected layers. The classification
accuracy results are shown in Table \ref{table:cifar} and demonstrate that the
SAE approach improves the accuracy of both shallow and deep networks, with the
most significant improvement observed in the shallow CIFAR-10 network (from
56\% to 68\% accuracy). \ Using the Average-Pooling based SAE also improved
the baseline results significantly, implying that using additional input image
augmentations carried the heavy load in the SAE.\begin{table}[tbh]
\caption{\textbf{Self-Attention-based Embedding.} The SAE was applied to the
CIFAR-10 and CIFAR-100 datasets using different backbones. AvgPull refers to
implementing the SAE using Average Pooling instead of a Transformer-Encoder.}%
\label{table:cifar}%
\centering
\renewcommand{\arraystretch}{1.3}
\begin{tabular}
[c]{@{}lccccc}%
\toprule\textbf{Dataset} & \textbf{Backbone} & \textbf{\#Aug} &
\textbf{\#layer} & \textbf{\#heads} & \textbf{Acc. [\%]}\\
\midrule CIFAR-10 & Shallow & 1 &  &  & 56\\
CIFAR-10 & Shallow & 5 & \multicolumn{2}{c}{AvgPull} & \textbf{72}\\
CIFAR-10 & Shallow & 3 & 1 & 2 & 62\\
CIFAR-10 & Shallow & 5 & 1 & 2 & 70\\
CIFAR-10 & Shallow & 5 & 2 & 2 & \textbf{72}\\
CIFAR-10 & Shallow & 10 & 1 & 2 & \textbf{72}\\
CIFAR-10 & Shallow & 15 & 1 & 2 & \textbf{72}\\
\midrule CIFAR-10 & VGG16 & 1 &  &  & 87\\
CIFAR-10 & VGG16 & 5 & \multicolumn{2}{c}{AvgPull} & 90\\
CIFAR-10 & VGG16 & 5 & 1 & 2 & \textbf{90}\\
CIFAR-10 & VGG16 & 5 & 2 & 2 & \textbf{92}\\
CIFAR-10 & VGG16 & 10 & 1 & 2 & \textbf{90}\\
\specialrule{1pt}{1\jot}{0pc}CIFAR-100 & VGG16 & 1 &  &  & 60\\
CIFAR-100 & VGG16 & 5 & 1 & 2 & 63\\
CIFAR-100 & VGG16 & 10 & 1 & 2 & \textbf{65}\\
\midrule CIFAR-100 & RESNET18 & 1 &  &  & 59\\
CIFAR-100 & RESNET18 & 5 & \multicolumn{2}{c}{AvgPull} & 64\\
CIFAR-100 & RESNET18 & 5 & 1 & 2 & \textbf{66}\\
CIFAR-100 & RESNET18 & 10 & 1 & 2 & \textbf{67}\\
\midrule CIFAR-100 & RESNET34 & 1 &  &  & 61\\
CIFAR-100 & RESNET34 & 5 & \multicolumn{2}{c}{AvgPull} & 66\\
CIFAR-100 & RESNET34 & 5 & 1 & 2 & \textbf{67}\\
CIFAR-100 & RESNET34 & 5 & 2 & 2 & \textbf{67}\\
CIFAR-100 & RESNET34 & 10 & 1 & 2 & \textbf{67}\\
\midrule CIFAR-100 & RESNET50 & 1 &  &  & 58\\
CIFAR-100 & RESNET50 & 5 & \multicolumn{2}{c}{AvgPull} & 65\\
CIFAR-100 & RESNET50 & 5 & 1 & 2 & \textbf{68}\\
CIFAR-100 & RESNET50 & 10 & 1 & 2 & \textbf{68}\\
\bottomrule &  &  &  &  &
\end{tabular}
\end{table}

\section{Conclusions}

In this paper, we present a novel approach for deep-learning age estimation
from face images. First, we show that multiple augmentations can be encoded
and aggregated by a Transformer-Encoder to yield a robust embedding. The
second step is to propose a hierarchical probabilistic age estimation
framework based on a deep classifier for estimating age probabilities.
Probability estimates are used to weight a corresponding ensemble of local
regressors, each adapted to a particular age subdomain. Our proposed scheme
outperforms the current SOTA schemes. We also present, for the first time to
our knowledge, a bias analysis of the SOTA results in face-based age
estimation. We hope that the proposed bias analysis will be used by others in
the field.

\balance
{\small
\bibliographystyle{IEEEtran}
\bibliography{references}
}

\end{document}